\pdfoutput=1
\documentclass[11pt]{article}

\usepackage{fullpage,times}
\usepackage{multirow}
\usepackage{parskip}
\usepackage{wrapfig}

\newcommand{\ours}{{\bf DiversePatch (ours)}}

\usepackage{amsmath, amsthm, amssymb, amsfonts, mathtools, graphicx, enumerate}
\usepackage{adjustbox}
\usepackage[utf8]{inputenc} 
\usepackage[T1]{fontenc}    
\usepackage{url}            
\usepackage{booktabs}
\usepackage{nicefrac}       
\usepackage{microtype}      
\usepackage{xspace}
\usepackage{algorithm,algorithmic}
\usepackage{color}
\usepackage{enumitem}
\usepackage{comment}
\usepackage{bm}
\usepackage{apptools}
\usepackage[page, header]{appendix}
\usepackage{titletoc}
\usepackage{array}

\usepackage[scaled=.9]{helvet}

\usepackage{url}
\usepackage{caption}
\usepackage[table,dvipsnames]{xcolor}

\usepackage{natbib}
\bibpunct{(}{)}{;}{a}{,}{,}

\usepackage{amsthm}   
\begingroup
    \makeatletter
    \@for\theoremstyle:=definition,remark,plain\do{%
        \expandafter\g@addto@macro\csname th@\theoremstyle\endcsname{%
            \addtolength\thm@preskip\parskip
            }%
        }
\endgroup

\newcommand{\todo}[1]{\textcolor{gray}{TODO: #1}}
\newcommand{\red}[1]{\textcolor{red}{#1}}
\newcommand{\med}[1]{\textcolor{magenta}{#1}}

\newcommand{\myempty}[1]{}

\newcommand{\vx}{\boldsymbol{x}}
\newcommand{\vv}[1]{\boldsymbol{#1}}

\newcommand{\yrm}[1]{}

\newcommand{\qiangremoved}[1]{}

 \def\bb#1\ee{\begin{align*}#1\end{align*}}

 \def\bba#1\eea{\begin{align}#1\end{align}}

\usepackage{lipsum}  
\makeatletter
\newcommand{\printfnsymbol}[1]{%
  \textsuperscript{\@fnsymbol{#1}}%
}
\makeatother

\title{Vision Transformers with Patch Diversification}

\author{
~~Chengyue Gong\textsuperscript{1},~~Dilin Wang~\textsuperscript{2}, ~~Meng Li\textsuperscript{2},~~Vikas Chandra\textsuperscript{2},~~Qiang Liu\textsuperscript{1} \\
\textsuperscript{1} University of Texas at Austin \hspace{10pt} \textsuperscript{2} Facebook \\
{\tt \small \{cygong, lqiang\}@cs.utexas.edu, \{wdilin, meng.li, vchandra\}@fb.com}
}
\date{}

\begin{document}

\maketitle

\vspace{-1.5em}

\begin{abstract}
Vision transformer has demonstrated promising performance on challenging computer vision tasks. However, directly training the vision transformers may yield unstable and sub-optimal results. Recent works propose to improve the performance of the vision transformers by modifying the transformer structures, e.g., incorporating convolution layers. In contrast, we investigate an orthogonal approach to stabilize the vision transformer training without modifying the networks. We observe the instability of the training can be attributed to the significant similarity across the extracted patch representations. More specifically, for deep vision transformers, the self-attention blocks tend to map different patches into similar latent representations, yielding information loss and performance degradation. To alleviate this problem, in this work, we introduce novel loss functions in vision transformer training to explicitly encourage diversity across patch representations for more discriminative feature extraction. We empirically show that our proposed techniques stabilize the training and allow us to train wider and deeper vision transformers. We further show the diversified features significantly benefit the downstream tasks in transfer learning. For semantic segmentation, we enhance the state-of-the-art (SOTA) results on Cityscapes and ADE20k to 83.6 and 54.5 mIoU, respectively. Our code is available at \url{https://github.com/ChengyueGongR/PatchVisionTransformer}.
\end{abstract}

\section{Introduction}

    
    
    

Recently, vision transformers have demonstrated promising performance on various challenging computer vision tasks, including image classification~\citep{dosovitskiy2020vit}, object detection~\citep{zhu2020deformable, carion2020end}, multi-object tracking~\citep{meinhardt2021trackformer}, image generation~\citep{jiang2021transgan} and video understanding~\citep{bertasius2021space}. 
Compared with highly optimized convolutional neural networks (CNNs), e.g., ResNet~\citep{he2016deep} and  EfficientNet~\citep{tan2019efficientnet}, transformer encourages non-local computation and achieves comparable, and even better performance when pre-trained on large scale datasets.

For a vision transformer, an image is usually split into patches and the sequence of linear embeddings of these patches are provided as the input to the stacked transformer blocks~\citep{dosovitskiy2020vit}. A vision transformer can learn the patch representations effectively using the self-attention block, which aggregates the information across the patches~\citep{vaswani2017attention}. The learned patch representations are then used for various vision tasks, such as image classification, image segmentation, and object detection. Hence, learning high-quality and informative patch representations becomes key to the success of a vision transformer.

Though promising results on vision tasks have been demonstrated for vision transformers, it is found that the training of vision transformers is not very stable, especially when the model becomes wider and deeper~\citep{touvron2021going}. To understand the origin of the training instability, we use two popular vision transformer variants, i.e., DeiT~\citep{touvron2020training} and SWIN-Transformer~\citep{liu2021swin}, and study the extracted patch representations of each self-attention layer. We find the average patch-wise absolute cosine similarity between patch representations increases significantly for late layers in both two models. For a 24-layer DeiT-Base model, the cosine similarity can reach more than 0.7 after the last layer. This indicates a high correlation and duplication among the learned patch representations. Such behavior is undesired as it degrades overall representation power of patch representations and reduces the learning capacity of powerful vision transformers. Our findings shed light on the empirical results found by \citep{touvron2021going}, and may partially explain the reason why simply increasing the depth of standard vision transformers cannot boost the model performance.

To alleviate the problem in vision transformers, we propose three different techniques. First, We propose to directly promote the diversity among different patch representations by penalizing the patch-wise cosine similarity. Meanwhile, we observe the input patch representations to the first self-attention layer are often more diversified as they rely solely on the input pixels. Based on this observation, we propose a patch-wise contrastive loss to encourage the learned representations of the same patch to be similar between the first and layer layers while force the representations of different patches in one given image to be different.


Thirdly, we propose a patch-wise mixing loss. Similar to Cutmix~\citep{yun2019cutmix}, we mix the input patches from two different images and use the learned patch representations from each image to predict its corresponding class labels. With the patch-wise mixing loss, we are forcing the self-attention layers to only attend to patches that are most relevant to its own category, and hence, learning more discriminating features.


Empirically, leveraging the proposed diversity-encouraging training techniques, 
we significantly improve the image classification for standard vision transformers on ImageNet without any architecture modifications or any extra data. 
Specifically, we achieve 83.3\% top-1 accuracy on ImageNet with an input resolution of 224$\times$224 for DeiT.
We also finetune the checkpoint trained on ImageNet-22K from SWIN-Transformer \citep{liu2021swin} and achieve 87.4\% top-1 accuracy on ImageNet.
By transferring the backbone model to semantic segmentation, we enhance the SOTA results on Cityscapes and ADE20k valid set to 83.6 and 54.5 mIoU, respectively.



\section{Preliminaries: Vision Transformers}

In this section, we give a brief introduction 
to vision transformers~\citep{dosovitskiy2020vit}.
Given a training example $(\bm{x}, y)$, where $\bm x$ and $y$ denote the input image and the label, respectively. 
A vision transformer first splits $\bm x$ into a set of non-overlapping patches, i.e., $\bm{x} = (x_1, \cdots, x_n)$, 
where $x_i$ denotes a patch and $n$ is the total number of patches. 
Each patch $x_i$ is then transformed into a latent representation via a projection layer, and augmented with a position embedding. 
A learnable \emph{class} patch is introduced to capture the label information. 
During training, a vision transformer model gradually transforms patch representations  $\bm{h}^{[\ell]} = (h_{class}^{[\ell]}, h_1^{[\ell]}, \cdots, h_n^{[\ell]})$ with a stack of self-attention layers~\citep{vaswani2017attention}. 
Here, $\ell$ denotes the layer index and $h_{class}^{[\ell]}$ and $h_i^{[\ell]}$ denote the learned \emph{class} patch and the image patch of $x_i$ at the $\ell$-th layer, respectively.
Let $L$ denote the total number of layers and $g$ denote a classification head. The vision transformer is trained by minimizing a classification loss $\mathcal{L}(g(h_{class}^{[L]}), y)$. See Figure~\ref{fig:oversmooth} (a) for an overview of vision transformers. It has been reported that the training of vision transformer can suffer from instability issue, especially for deep models \citep{touvron2020training}.

\section{Examining Patch Diversity in Vision Transformers}
\label{sec:study}

To understand the origin of the training instability, we study the patch representations learned after each self-attention layer. Ideally, we would hope the patch representations to be diverse and capture different information from the inputs. We study the diversity of the patch representations by computing the patch-wise absolute cosine similarity. Consider a sequence of patch representations $\bm{h} = [h_{class}, h_1, \cdots, h_n]$. We define the patch-wise absolute cosine similarity as
\begin{equation*}
\mathcal{P}(\bm h) = \frac{1}{n(n-1)} \sum_{i\neq j} \frac{|h_i^\top h_j|}{\parallel h_i \parallel_2 \parallel h_j \parallel_2}.
\end{equation*}
Here we ignore the \emph{class} patch. Larger values of  $\mathcal{P}(\bm h)$ indicate a higher correlation among patches and vice versa.

We test two variants of vision transformers pre-trained on the ImageNet dataset~\citep{deng2009imagenet}, including a 24-layer DeiT-Base model~\citep{touvron2020training} (denoted as DeiT-Base24) and a SWIN-Base~\citep{liu2021swin} model. We also evaluate a ResNet-50~\citep{he2016deep} model for comparison. For the DeiT-Base24 model, we uniformly select 5 layers to compute the patch-wise cosine similarity. For the SWIN-Base model and ResNet-50 model, we select the input representation to each down-sampling layer. Specifically for the ResNet-50 model, we regard the representations at each spatial location as an individual patch.

\begin{figure}[t]
\centering
\setlength{\tabcolsep}{1pt}
\begin{tabular}{cc}
\includegraphics[height=0.225\textwidth]{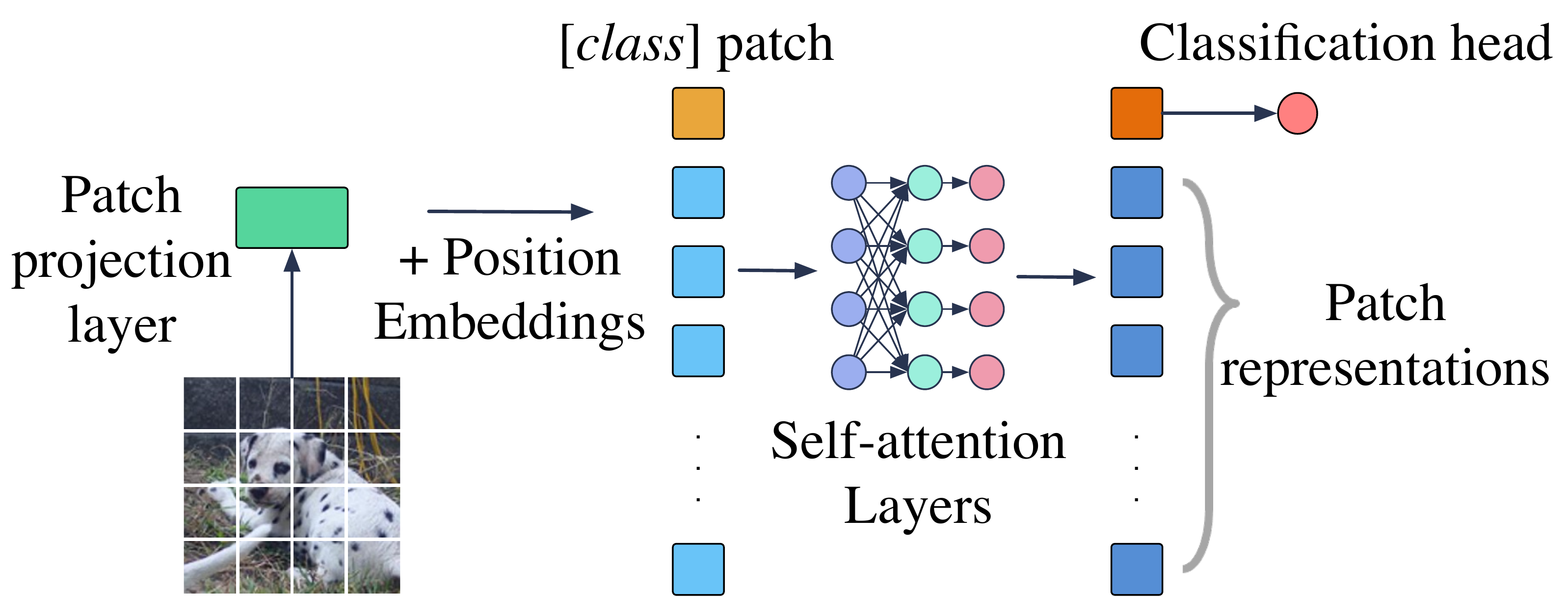} &
\raisebox{1.2em}{\rotatebox{90}{\small Cosine similarity}}
\includegraphics[height=0.225\textwidth]{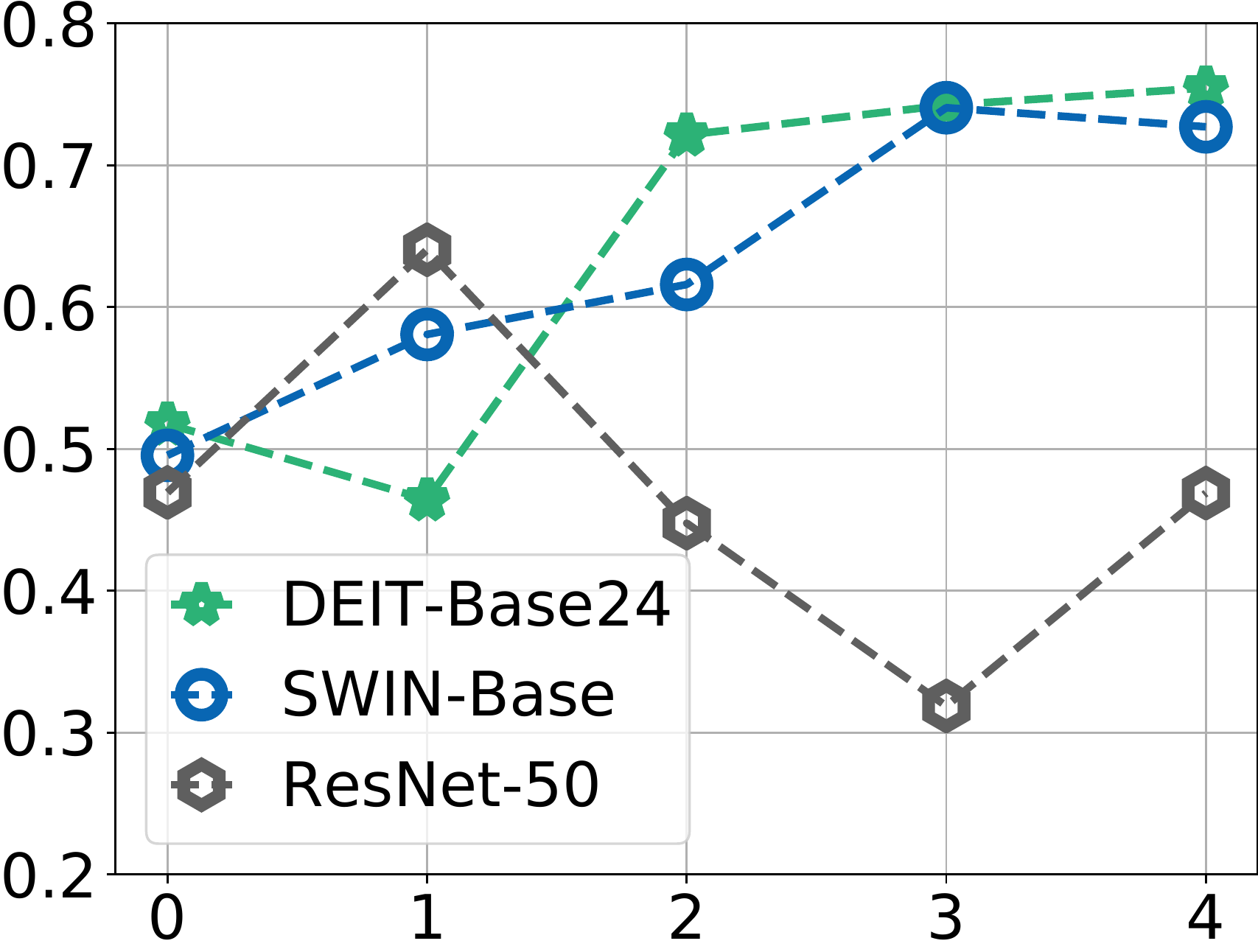}\\ 
& {\small  Block index}   \\
(a) An illustration of vision transformers 
& (b) Pairwise absolute cosine similarity \\
\end{tabular}
\caption{(a) An overview of vision transformers by following \citep{dosovitskiy2020vit}. Each image patch is first transformed to a latent representation using a linear patch projection layer. The \emph{dog} image is from ImageNet~\citep{deng2009imagenet}. 
(b) Comparisons of patch-wise absolute cosine similarities.
All similarities are computed with 10,000 sub-sampled images from the  ImageNet training set without data augmentation.}
\label{fig:oversmooth}
\end{figure}

As shown in Figure~\ref{fig:oversmooth} (b), the patch-wise cosine similarity $\mathcal{P}(\cdot)$ of the patch representations learned by DeiT-Base24 and SWIN-Base gradually increases with the the depth of layers. For DeiT-Base24, the average cosine similarity becomes larger than 0.7 for the representations after the last layer. In contrast, ResNet-50 learns relative diversified features across the network without an obvious increase of the patch-wise cosine similarity. Such high similarity across the learned patch representations is undesired as it degrades the learning capacity of the vision transformer models and limits the actual information extracted from the input images. There is also risk for the patch representations to degenerate with the increase of model depth, which may partially explain the high training instability of deep models.

\section{DiversePatch: Promoting Patch Diversification for Vision Transformers}
\label{sec:method}

To alleviate the observed problem, we propose three regularization techniques to encourage diversity across the learned patch representations.

\begin{figure}[ht]
\centering
\setlength{\tabcolsep}{2pt}
\begin{tabular}{ccc}
\includegraphics[height=0.19\textwidth]{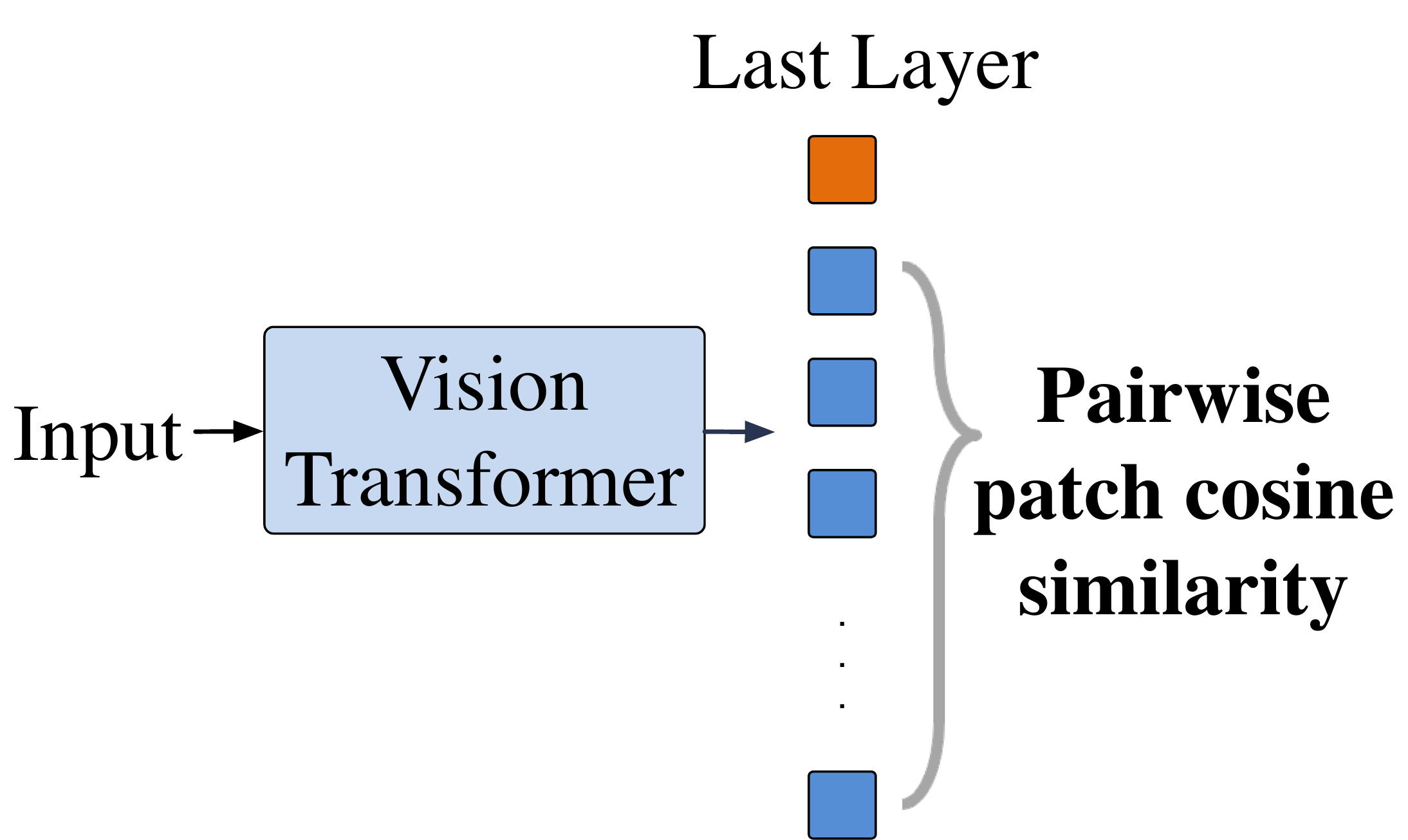} &
\includegraphics[height=0.19\textwidth]{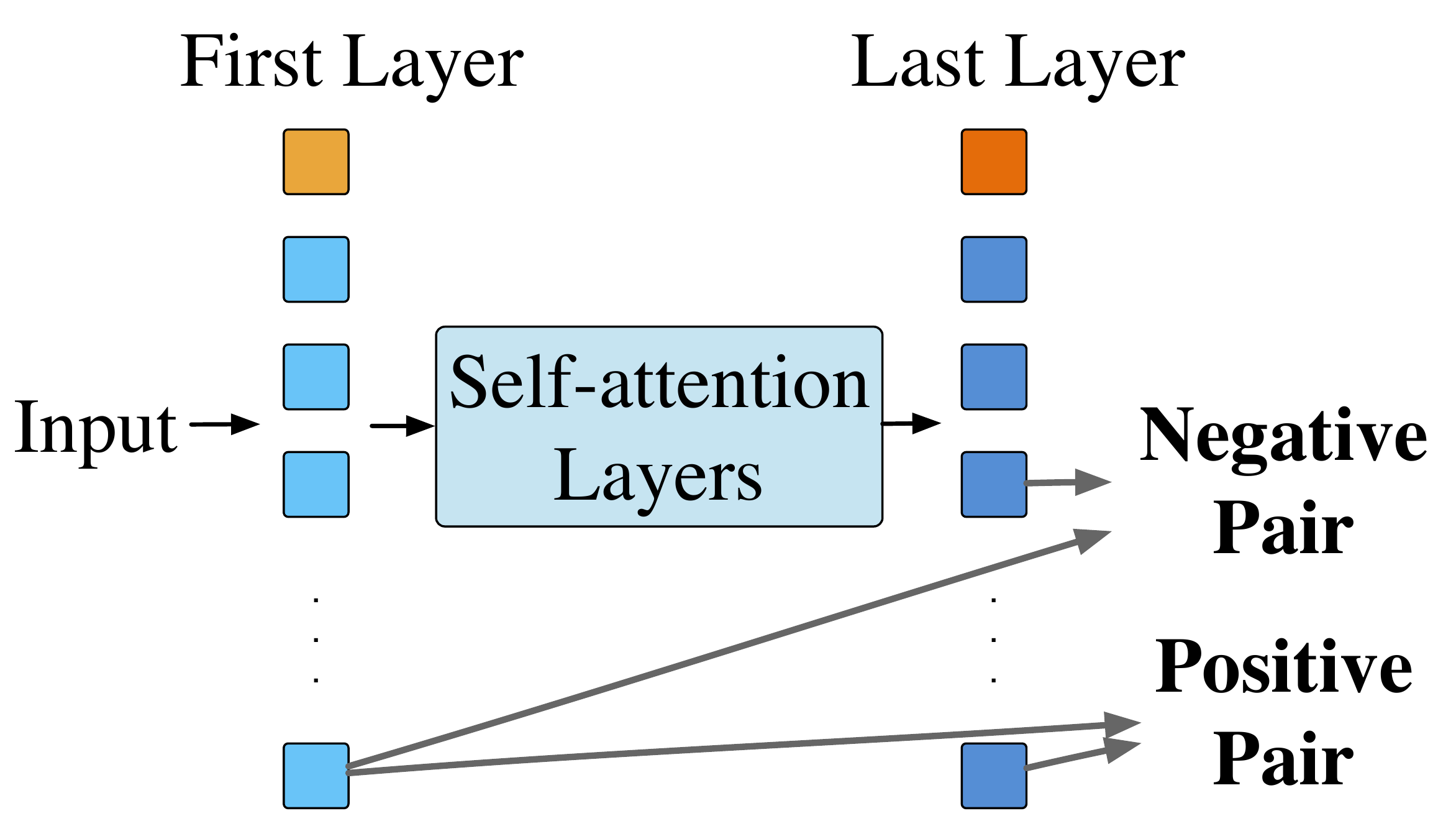} &
\includegraphics[height=0.19\textwidth]{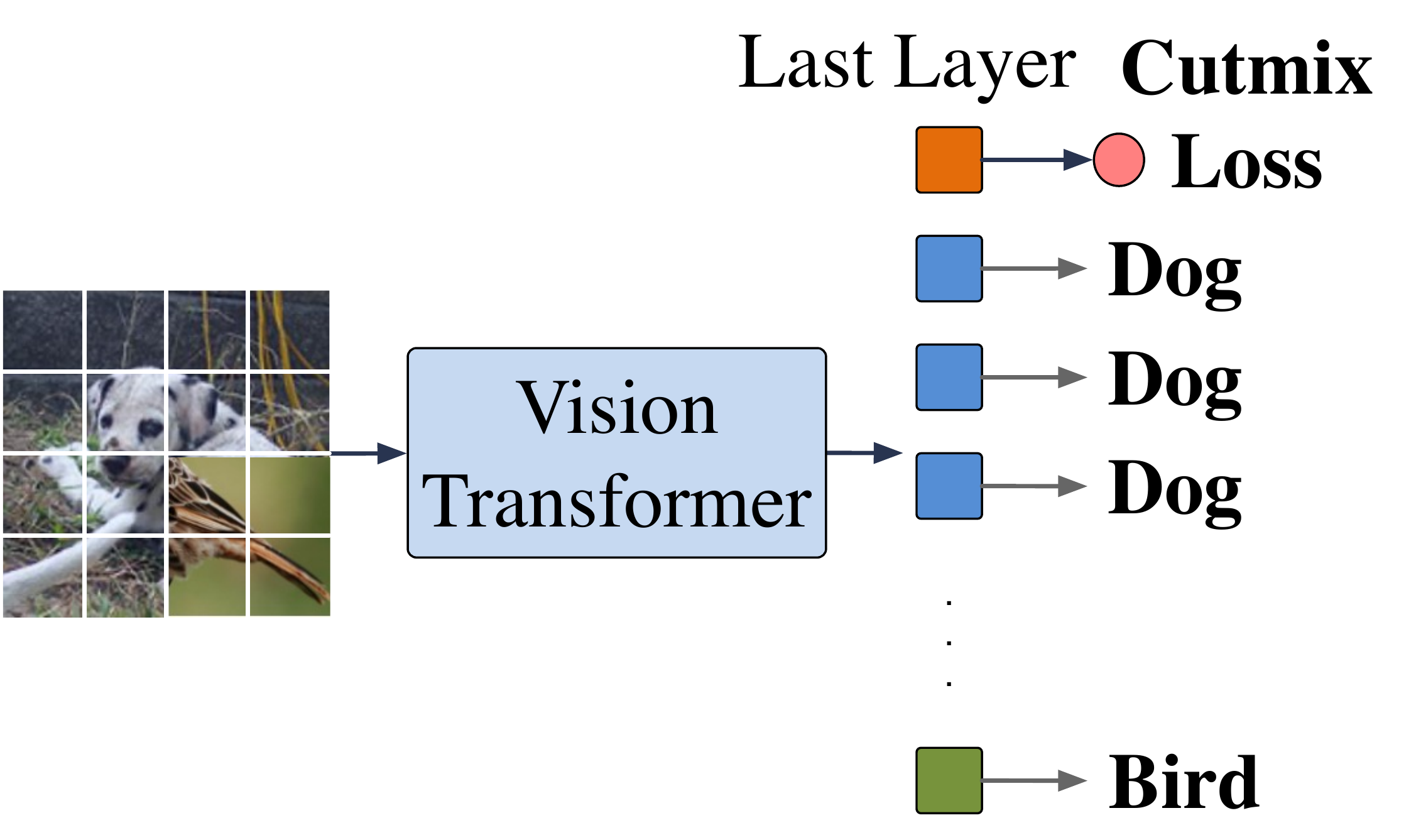} \\
(a) Patch-wise Cosine Loss & (b) Patch-wise Contrastive Loss  & 
(c) Patch-wise Mixing Loss \\
\end{tabular}
\caption{ An illustration of our patch diversification promoting losses. (a) Patch-wise cosine loss. (b) Patch-wise contrastive loss. (c) Patch-wise mixing loss.
}
\label{fig:demo}
\end{figure}

\paragraph{Patch-wise cosine loss}
As a straightforward solution, we propose to directly minimize the patch-wise absolute value of cosine similarity between different patch representations. 
Given the final-layer patch representations $\bm{h}^{[L]}$ of an input $\bm x$,
we add a patch-wise cosine loss to the training objective:
\begin{equation*}
\mathcal{L}_{\textit{cos}}(\bm x) = \mathcal{P}(\bm{h}^{[L]}).
\end{equation*}
This regularization loss explicitly minimizes the pairwise cosine similarity between different patches.
Similar approaches have also been adopted for training improved 
representations in graph neural networks \citep{chen2020measuring} and diversified word embeddings in language models \citep{gao2019representation}.
Meanwhile, 
this regularization loss can be viewed as minimizing the upper bound of the largest eigen-value of $\bm{h}$ \citep{merikoski1984trace}, hence improving the expressiveness of the representations. See Figure~\ref{fig:demo} (a) for an illustration.

\paragraph{Patch-wise contrastive loss}
Secondly, as shown in Figure~\ref{fig:oversmooth} (b), 
representations learned in early layers are more diverse compared to that in deeper layers.
Hence, we propose a contrastive loss which uses the representations in early layers, and regularizes the patches in deeper layers to reduce the similarity of patch representations. 
Specifically,  given an image $\bm{x}$, 
let $\bm{h}^{[1]}=\{h_i^{[1]}\}_i$ and  $\bm{h}^{[L]}=\{h_i^{[L]}\}_i$ denote its patches at the first and the last layer, respectively.
We constrain each $h^{[L]}_i$ to be similar to $h^{[1]}_i$ and to be different to any other patches $h^{[1]}_{j\neq i}$ as follows (see Figure~\ref{fig:demo} (b)),
\begin{equation*}
\mathcal{L}_{\textit{contrastive}}(\bm x)= - \frac{1}{n} \sum_{i=1}^n \log \frac{\exp({h^{[1]}_i}^\top h^{[L]}_i)}{\exp({h^{[1]}_i}^\top h^{[L]}_i) + \exp({h^{[1]}_i}^\top (\frac{1}{n-1}\sum_{j \neq i} h^{[L]}_j))}, 
\end{equation*}
In practice, we stop the gradient on $\bm{h}^{[1]}$.

\paragraph{Patch-wise mixing loss}
Thirdly, instead of just using the class patch for the final prediction, we propose to train each patch to predict the class label as well. This can be combined with Cutmix~\citep{yun2019cutmix} data augmentation to provide additional training signals for the vision transformer. As shown in Figure~\ref{fig:demo} (c), we mix the input patches from two different images and attach an additional shared linear classification head to each output patch representations for classification. The mixing loss forces each patch to only attend to its a subset of patches from the same input images and ignore unrelated patches. Hence, it effectively prevents simple averaging across different patches to yield more informative and useful patch representations. This patch mixing loss can be formulated as below,
\begin{equation*}
\mathcal{L}_\textit{mixing}(\bm x) = \frac{1}{n} \sum_{i=1}^n \mathcal{L}_\textit{ce}(g(h^{[L]}_i), y_i),
\end{equation*}
where $h^{[L]}_i$ represents patch representations in the last layer, $g$ denotes the additional linear classification head,  $y_i$ stands for the patch-wise class label and $\mathcal{L}_\textit{ce}$ denotes the cross entropy loss.

\paragraph{Algorithm}
We improve the training of vision transformers by simply jointly minimizing the weighted combination of $\alpha_1 \mathcal{L}_{\textit{cos}} + \alpha_2 \mathcal{L}_{\textit{contrastive}} +\alpha_3 \mathcal{L}_{\textit{mixing}}$. 
Our method does not require any network modifications, and in the meantime, is not restricted to any specific architectures.
In our experiments, we simply set $\alpha_1=\alpha_2=\alpha_3=1$ without any particular hyper-parameters tuning.

\section{Experiments}
\label{sec:experiment}

In this section, we apply our method to improve the training of a variety of 
vision transformers, including DeiT \citep{touvron2020training} and SWIN-Transformer \citep{liu2021swin}, and evaluate on a number of image classification and semantic segmentation benchmarks. 
We show that our training strategy promotes patch diversification and 
learns transformers with significantly better transfer learning performance on downstream semantic segmentation tasks.

\begin{table}[t]
    \centering
    \setlength{\tabcolsep}{2pt}
    \begin{tabular}{l|l|c|c|c|c}
    \hline 
    & Method & \# Params (M) & Input Size & $+$ Conv & Top-1 (\%) \\ 
    \hline
    \hline
    \multirow{6}{*}{CNNs} 
    & ResNet-152 \scriptsize{\citep{he2016deep}} &  230 & 224 &$\checkmark$ & 78.1 \\
    & DenseNet-201 \scriptsize{\citep{huang2017densely}} & 77 & 224 & $\checkmark$ & 77.6 \\
    & EffNet-B8 \scriptsize{\citep{gong2020maxup}} & 87 & 672 &$\checkmark$  & 85.8 \\
    & EffNetV2-L \scriptsize{\citep{tan2021efficientnetv2}} & 121 & 384 &$\checkmark$  & 85.7 \\
    &NFNet \scriptsize{\citep{brock2021high}} & 438 & 576 &$\checkmark$  & 86.5 \\
    & LambdaNet-420 \scriptsize{\citep{bello2021lambdanetworks}} & 87 & 320 &$\checkmark$  & 84.9 \\
    \hline
    \hline
    \multirow{2}{*}{CNNs + } 
    & CVT-21 \scriptsize{\citep{wu2021cvt}} & 32 & 224 & $\checkmark$ & 82.5 \\
    \multirow{2}{*}{Transformers} 
    & CVT-21 & 32 & 384 & $\checkmark$ & 83.3 \\
    \cline{2-6}
    & LV-ViT-M \scriptsize{\citep{jiang2021token}} & 56 & 224 & $\checkmark$ & 84.0 \\
    & LV-ViT-L & 150 & 448 & $\checkmark$ & 85.3 \\
    \hline
    \hline
    \multirow{6}{*}{DeiT {(scratch)}} & DeiT-Small12 \scriptsize{\citep{touvron2020training}} & \multirow{ 2}{*}{22} & \multirow{ 2}{*}{224} & \multirow{ 2}{*}{$\times$} & 80.4 \\
    & + \bf{\ours} & & & & \bf{81.2} \\
    \cline{2-6}
    & DeiT-Small24  & \multirow{ 2}{*}{44} & \multirow{ 2}{*}{224} & \multirow{ 2}{*}{$\times$} & 80.3 \\
    & + \ours  & & & &  \bf{82.2} \\
    \cline{2-6}
    & DeiT-Base12 &  \multirow{ 2}{*}{86} & \multirow{ 2}{*}{224} & \multirow{ 2}{*}{$\times$} & 82.1 \\
    & + \ours  & & & &  \bf{82.9} \\
    \cline{2-6}
    & DeiT-Base24 &  \multirow{2}{*}{172} & \multirow{ 2}{*}{224} & \multirow{2}{*}{$\times$} & 82.1 \\
    & + \ours& & & & \bf{83.3} \\
    \cline{2-6}
    & DeiT-Base12 & \multirow{2}{*}{86} & \multirow{ 2}{*}{384} & \multirow{2}{*}{$\times$} & 83.6 \\
    & + \ours & & & & \bf{84.2} \\
    \hline
    \hline
    \multirow{4}{*}{SWIN {(finetune)}} & SWIN-Base \scriptsize{\citep{liu2021swin}} & \multirow{2}{*}{88} & \multirow{ 2}{*}{224} & \multirow{2}{*}{$\times$} & 83.4 \\
    & +\ours & & & & \bf{83.7} \\
    \cline{2-6}
    & SWIN-Base & \multirow{2}{*}{86} & \multirow{ 2}{*}{384} & \multirow{2}{*}{$\times$} & 84.5 \\ 
    &  + \ours & & & & \bf{84.7} \\ 
    \hline
    \hline
    \end{tabular}
    \caption{Top-1 accuracy results on ImageNet. We train all DeiT based models from scratch for 400 epochs. For SWIN-Transformer based models, we finetune from existing checkpoints for 30 epochs.
    Results without any extra data are reported.
    }
    \label{tab:main_result}
\end{table}

\subsection{Main Results on ImageNet}
\label{sec:exp_imagenet}

We use DeiT~\citep{touvron2020training} and SWIN transformers~\citep{liu2021swin} as our baseline models, and improving these models by incorporating our patch diversification-promoting losses to the training procedure.

\paragraph{Settings}
For DeiT based models, 
we closely follow the training settings provided in \citet{touvron2020training}~\footnote{\url{https://github.com/facebookresearch/deit}}.
We train all DeiT baselines and our models for 400 epochs. 
We use stochastic depth dropout and linearly increase the depth dropout ratio from 0 to .5 following ~\citet{huang2016deep}. 
Additionally,  
we use stochastic depth dropout ratio of 0.5 for DeiT-Base24 and DeiT-Small24, 
which allows us to train deeper DeiT models without diverging. 
For our method, we remove \emph{MixUp} \citep{zhang2017mixup} 
and repeated data augmentation \citep{hoffer2020augment} as they are not compatible with our \emph{path-wise mixing loss}. Detailed ablation study is in Section \ref{sec:ablation}.  


For SWIN transformers, we use the official code for training, evaluation and finetuning \footnote{\url{https://github.com/microsoft/Swin-Transformer}}.
Specifically, we download the official SWIN-Base models pretrained with resolution of $224\times 224$ and $384\times 384$, respectively.
We further finetune them for another 30 epochs with or without our patch diversification losses.
In particular, we use a batch size of 1024 (128 X 8 GPUS), a constant learning rate of $10^{-5}$ and a weight decay of $10^{-8}$ for finetuning.


\paragraph{Results on ImageNet}

As demonstrated in Table \ref{tab:main_result}, for all the model architectures we evaluated, 
our method leads to consistent improvements upon its corresponding baseline model. For example, for DeiT based models, 
we improve the top-1 accuracy of DeiT-Small12 from 80.4\% to 81.2\%,
and improve the top-1 accuracy of DeiT-Base12 from 82.1\% to 82.9\%.
A similar trend can also been for SWIN based models.

Additionally, our method allows us to train more accurate vision transformers by simply stacking more layers. 
Naive training of DeiT models cannot benefit from larger and deeper models due to rapid patch representation degradation throughout the networks. 
As we can see from Table~\ref{tab:main_result}, with standard training, the results from DeiT-Small24 and DeiT-Base24 are not better than the results from DeiT-Small12 and DeiT-Base12;
on the other hand, our method alleviates over-smoothing and learns more  diversified patch features, enabling further improvements to the large model regime. Specifically, our method achieves 82.2\%  and 83.3\% top-1 accuracy for DeiT-Small24 and DeiT-Base24, respectively, which are 
$1.9\%$ and $1.2\%$ higher compared to their corresponding DeiT baselines.

To further verify reproducibility of our results, we run DeiT-Base12 for three trials,
achieving top-1 accuracy of 82.86\%, 82.82\%, 82.89\%; the round to round performance variant is negligible and the \emph{s.t.d} is smaller than 0.1.

\begin{wrapfigure}{r}{0.5\textwidth}
\centering
\vspace{-1.2em}
\begin{tabular}{c}
\raisebox{1.1em}{\rotatebox{90}{{Avg. Cosine similarity}}} 
\includegraphics[width=0.45\textwidth]{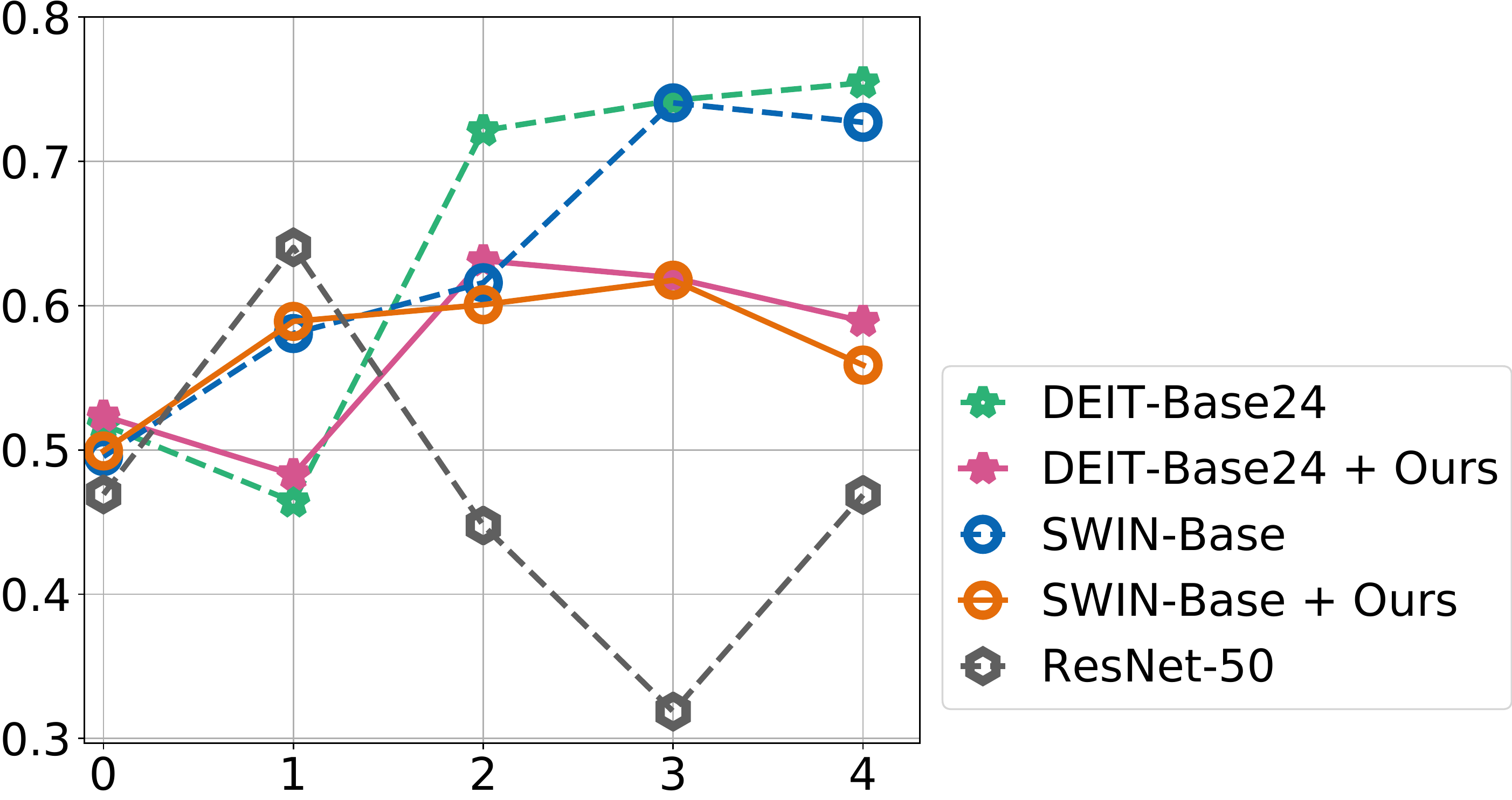} \\
{{Block Index~~~~~~~~~~~~~~~~~~~~~}}  \\ 
\end{tabular}
\vspace{-1.2em}
\caption{Comparison on average patch-wise absolute cosine similarity.
}
\vspace{-1.2em}
\label{fig:exp_ours_cosine_sim}
\end{wrapfigure}
Following the studies in Section~\ref{sec:study}, 
we plot the patch-wise absolute cosine similarity for the patch features learned by our method in Figure~\ref{fig:exp_ours_cosine_sim}. As shown in Figure~\ref{fig:exp_ours_cosine_sim}, 
the patch-wise absolute cosine similarity is reduced significantly for both DeiT-Base24 and SWIN-Base;
the cosine similarity among the learned patches 
in the last layer is similar to the result from the ResNet-50 baseline.



We also compare with recently proposed CNN and transformer hybrids in Table~\ref{tab:main_result},
including CVT \citep{wu2021cvt} and LV-ViT \citep{jiang2021token}.   
These hybrid models introduce additional convolution layers to both the patch projection layers and self-attention layers, and achieve better classification accuracy on ImageNet compared to pure transformers like DeiT and SWIN-transformers. 
These hybrid approaches are non-conflict to our method, and we will further study the diverse patch representations in these architectures and apply our method to them.

\paragraph{Results with ImageNet-22K Pre-training}

We also fine-tune ViT-Large models \citep{dosovitskiy2020vit} and SWIN-Large models that are pretrained on ImageNet-22K \citep{russakovsky2015imagenet} to further push the limits of accuracy on ImageNet.
ImageNet-22k contains 22k classes and  14M images. 
Specifically, we directly download the ImageNet-22K pre-trained models provided in ViT and SWIN-Transformer and finetune 
these checkpoints on ImageNet training set with 30 epochs, 
a batch size of 1024, a constant learning rate of $10^{-5}$ and a weight decay of $10^{-8}$.

Table~\ref{tab:imagenet22k} shows the finetuning accuracy on ImageNet.
Our method again leads consistent improvements across all evaluated settings. Specifically, we improve the VIT-Large top-1 accuracy from 85.1\% to 85.3\% and archive 87.4\% top-1 accuracy with SWIN-Large. 
As a future work, we will further pretrain the models on ImageNet-22k with our method to see whether the improvement is larger.

\begin{table}[ht]
    \centering
    \setlength{\tabcolsep}{6pt}
    \begin{tabular}{l|c|c}
    \hline 
    Model & Input Size & Top-1 Acc (\%)\\  
    \hline 
    VIT-Large \scriptsize{\citep{dosovitskiy2020vit}} & 224 & 83.6 \\
    + \ours & 224 & \bf{83.9}\\
    \hline
    VIT-Large  & 384 & 85.1\\
    + \ours & 384 & \bf{85.3}\\
    \hline
    SWIN-Large + ImageNet22k & 384 & 87.3 \\ 
    + \ours & 384 & \bf{87.4} \\
    \hline
    \end{tabular}
    \caption{
    Results on ImageNet by finetuing from  ImageNet-22K pretrained vision transformers. 
    }
    \label{tab:imagenet22k}
\end{table}

\subsection{Transfer Learning on Semantic Segmentation}
\label{sec:exp_segment}
Semantic segmentation requires a backbone network to extract representative and diversified features from the inputs; the downstream semantic segmentation performance critically relies on the quality of the extracted features. 
In this section, we use SWIN-Base pretrained on ImageNet (see Table~\ref{tab:main_result}) and SWIN-large pretrained on both ImageNet22k and ImageNet (see Table~\ref{tab:imagenet22k}) as our backbone models and finetune them on two widely-used semantic segmentation datasets,  ADE20K \citep{zhou2017scene} and Cityscapes \citep{cordts2016cityscapes}, to verify the transferability our pretrained models.

In particular, we show the backbone models trained with our diversity-promoting losses are especially helpful for downstream segmentation tasks. By using our pretrained SWIN-Large ($87.4\%$ top-1 accuracy),  our results outperform all existing methods and  establish a new state-of-the-art on ADE20K and the Cityscapes validation dataset, achieving mIOU of {54.5\%} and {83.6\%} mIoU, respectively. 

\paragraph{Datasets}
ADE20K \citep{zhou2017scene} is a large scene parsing dataset, covering 150 object and stuff categories.  ADE-20K contains 20K images for training, 2K images for validation, and 3K images for test images.
Cityscapes \citep{cordts2016cityscapes} dataset labels 19 different categories (with an additional unknown class) and consists of 2975 training images, 500 validation images and 1525 testing images. 
We do evaluation on the ADE20K and Cityscapes validation set in this paper.

\paragraph{Baselines}
We introduce both state-of-the-art CNNs \citep[e.g.][]{xiao2018unified, zhang2020resnest, wang2020hr, bai2020multiscale} and recent proposed transformer models \citep[e.g.][]{liu2021swin, zheng2020rethinking, ranftl2021vision} as our baselines.

\paragraph{Settings} 
We closely follow the finetuning settings proposed in SWIN transformers~\citep{liu2021swin} \footnote{\url{https://github.com/SwinTransformer/Swin-Transformer-Semantic-Segmentation}}. 
Specifically, 
we use UperNet \citep{xiao2018unified} in \emph{mmsegmentation} \citep{mmseg2020} as our testbed.
During training, we use AdamW \citep{loshchilov2017decoupled} optimizer with a learning rate of $6\times 10^{-5}$ and a weight decay of 0.01.
We use a cosine learning rate decay and a linear learning rate warmup of 1,500 iterations. 
We finetune our models for 160K iterations and 80K iterations on ADE20K and Cityscapes training set, respectively.
We adopt the default data augmentation scheme in \emph{mmsegmentation} \citep{mmseg2020} \footnote{\url{https://github.com/open-mmlab/mmsegmentation}} and train with 
512$\times$512 and 769$\times$769 crop size for ADE20K and Cityscapes, respectively, following the default setting in \emph{mmsegmentation}.
Additionally, 
following SWIN transformers \citep{liu2021swin}, we use a stochastic depth dropout of 0.3 for the first 80\% of training iterations, and increase the dropout ratio to 0.5 for the last 20\% of training iterations.

Following SWIN transformers~\citep{liu2021swin},
we report both the mIoUs from the single-scale evaluation and the mIoUs from the multi-scale flipping evaluation \citep{zhang2018context}. 
Specifically, for multi-scale flipping testing, we enumerate testing scales of  \{0.5, 0.75, 1.0, 1.25, 1.5, 1.75\} and random horizontal flip by following common practices in the literature \citep[e.g.][]{zhang2020resnest, liu2021swin, zheng2020rethinking}. 
The images are evaluated at 2048$\times$1024 and 2049$\times$1025 for ADE20K and cityscapes, respectively.


\paragraph{Results}

We summarize our results in Table \ref{tab:segmentation_ade20k} and Table \ref{tab:segmentation_cityscape}.
Our method improves the training of SWIN-Transformers and learns backbones capable of extracting more diversified features from the inputs, 
leading to new state-of-the-art segmentation performance on both ADE-20K and Cityscapes. Our achieve 54.5\% mIoU on ADE-20K and 83.6\% mIoU on the Cityscapes validation set, outperforming all existing approaches.



\begin{table}[ht]
    \centering
    \begin{tabular}{l|c|cc}
    \hline 
    Model & \#Params (M) & mIoU (\%) & mIoU (ms+flip) (\%)\\  
    \hline
    OCNet \scriptsize{\citep{yuan2019object}} & 56 & 45.5 & \texttt{N/A} \\
    UperNet \scriptsize{\citep{xiao2018unified}} & 86 & 46.9 & \texttt{N/A} \\
    ResNeSt-200 + DeepLab V3 & 88 & \texttt{N/A} & 48.4 \\
    SETR-Base\scriptsize{\citep{zheng2020rethinking}} & 98 & \texttt{N/A} & 48.3 \\
    SETR-Large \scriptsize{\citep{zheng2020rethinking}} & 308 & \texttt{N/A} & 50.3 \\
    DPT-ViT-Hybrid \scriptsize{\citep{ranftl2021vision}} & 90 & \texttt{N/A} & 49.0 \\
    DPT-ViT-Large \scriptsize{\citep{ranftl2021vision}} & 307 & \texttt{N/A} & 47.6 \\
    \hline 
    \hline
    Swin-Base & 121 & 48.1 & 49.7 \\
    + \ours & 121  & \bf{48.4}\scriptsize{$\pm$0.2} & \bf{50.1}\scriptsize{$\pm$0.2} \\
    \hline
    Swin-Large & 234  & 52.0 & 53.5 \\
    + \ours & 234 & \bf{53.1}\scriptsize{$\pm$0.1} & \bf{54.5}\scriptsize{$\pm$0.1} \\
    \hline
    \end{tabular}
    \caption{State-of-the-art on ADE20K. 
    `ms+flip' refers to mutli-scale testing with flipping \citep{zhang2018context}, and `\#Params' denotes the number of parameters.
    }
    \label{tab:segmentation_ade20k}
\end{table}

\begin{table}[ht]
    \centering
    \begin{tabular}{l|c|cc}
    \hline 
    Model & \#Params (M)  & mIoU (\%) & mIoU (ms+flip) (\%)\\  
    \hline 
    OCNet \scriptsize{\citep{yuan2019object}} & 56 & 80.1 & \texttt{N/A} \\
    HRNetV2 + OCR  \scriptsize{\citep{wang2020hr}} & 70 & 81.6 & \texttt{N/A} \\
    Panoptic-DeepLab  \scriptsize{\citep{cheng2019panoptic}} & 44 &  80.5 & 81.5 \\
    Multiscale DEQ \scriptsize{\citep{bai2020multiscale}}& 71 & 80.3 & \texttt{N/A} \\
    ResNeSt-200 + DeepLab V3 & 121  & \texttt{N/A} & 82.7 \\
    SETR-Base\scriptsize{\citep{zheng2020rethinking}} & 98 & \texttt{N/A} & 78.1 \\
    SETR-Large \scriptsize{\citep{zheng2020rethinking}}  & 308 & \texttt{N/A} & 82.1 \\
    \hline
    \hline
    Swin-Base & 121 & 80.4 & 81.5 \\
    + \ours & 121  & \bf{80.8}\scriptsize{$\pm$0.1} & \bf{81.8}\scriptsize{$\pm$0.1} \\
    \hline
    Swin-Large & 234 & 82.3 & 83.1 \\
    + \ours & 234 & \bf{82.7}\scriptsize{$\pm$0.2} & \bf{83.6}\scriptsize{$\pm$0.1} \\
    \hline
    \end{tabular}
    \caption{ State-of-the-art on the Cityscapes validation set. 
    }
    \label{tab:segmentation_cityscape}
\end{table}

\newcolumntype{C}{>{\centering\arraybackslash}m{2.3cm}}
\begin{table}[ht]
    \centering
    \setlength{\tabcolsep}{4pt}
    \begin{tabular}{CCC|C|C}
    \hline 
    Patch-wise cosine loss  & Patch-wise contrastive loss & Patch-wise mixing loss & DeiT-Base24 (\%) & Swin-Base ~~ (\%) \\    
    \hline 
    $\times$ & $\times$ & $\times$ & 82.1 & 83.4 \\
    \hline 
    $\checkmark$ & $\times$ & $\times$ & 82.5 & 83.5\\
    $\times$ & $\checkmark$ & $\times$ & 82.6 & \texttt{N/A} \\
    $\times$  & $\times$ & $\checkmark$& 82.8 & 83.4\\
    \hline
    $\checkmark$  & $\times$ & $\checkmark$& 83.1 & \bf{83.7} \\
    $\checkmark$ & $\checkmark$ & $\times$ & 83.1 & \texttt{N/A}  \\
    $\times$ & $\checkmark$ & $\checkmark$ & \bf{83.3}  & \texttt{N/A}  \\
    $\checkmark$ & $\checkmark$ & $\checkmark$ & \bf{83.3} &  \texttt{N/A}  \\
    \hline
    \end{tabular}
    \caption{Ablating the impact of different combinations of our proposed patch diversification losses.
    }
    \label{tab:loss_ablation}
\end{table}


\subsection{Ablation Studies}
\label{sec:ablation}

\paragraph{On the efficacy of our regularization strategies}
Our method introduces three regularization terms to promote patch diversification. In this part, 
we use DeiT-Base24 and SWIN-Base as our baseline models and ablate the effectiveness of  our \emph{patch-wise cosine loss}, \emph{patch-wise contrastive loss} and \emph{patch-wise mixing loss} by enumerating all different combination of training strategies. 
We proceed by exactly following the training settings in section~\ref{sec:exp_imagenet}.
We summarize our results in Table~\ref{tab:loss_ablation}.
As we can see from Table~\ref{tab:loss_ablation}, 
all diversification-promoting losses are helpful and 
all combinations lead to improved top-1 accuracy on ImageNet.
Specifically, 
we improved DeiT-Base24 from 82.1\% to 83.3\% on top-1 accuracy by combing all three losses;
for SWIN-Base, we did not ablate the \emph{patch-wise contrastive loss}, because the number of patches was reduced throughout the network due to down-sampling. In this case, we boost the  top-1 accuracy from 83.5\% and 83.7\% by incorporating the \emph{patch-wise cosine loss} and \emph{patch-wise mixing loss} into the training procedure.
And the patch representations learned by our method are particularly useful in down-stream semantic segmentation tasks, as demonstrated in section~\ref{sec:exp_segment}.


\paragraph{On the stabilization of training}
Vision transformers are prone to overfitting, and training successful vision transformers often requires careful hyper-parameters tuning. 
For example, DeiT uses a bag of tricks for stabilized training, including RandAugment~\citep{cubuk2020randaugment}, MixUp~\citep{zhang2017mixup}, CutMix~\citep{yun2019cutmix}, random erasing~\citep{zhong2020random},
stochastic depth~\citep{huang2016deep}, repeated augmentation~\citep{hoffer2020augment}, etc.
As we shown in Table~\ref{tab:ablate_deit_training}, 
removing some of these training tricks leads to significant performance degradation for DeiT. While our patch-wise diversification losses offer a natural regularization to prevent overfitting, and may therefore leading to a more stabilized training process. 
The models trained via our method yield consistently competitive results
across different training settings. 
Additionally, we show our method could further benefit from more advanced design of talking-heads attentions~\citep{shazeer2020talking}, longer training time and deeper architecture design, achieving consistent improvements upon our DeiT baselines. 



\begin{table}[ht]
    \centering
    \setlength{\tabcolsep}{12pt}
    \begin{tabular}{l|cc}
    \hline 
    Model & DeiT-Base12 & DeiT-Base12 +\ours \\    
    \hline 
    Standard (300 epochs) & 81.8 & 82.6 \\  
    + Talking Head & 81.8 & 82.7 \\
    \hline 
    - Repeat Augmentation & 78.4 & 82.7 \\
    - Random Erasing & 77.5 & 82.7 \\
    - Mixup & 80.3 & 82.7 \\
    - Drop Path & 78.8 & 80.2 \\ 
    \hline
    + 400 Epochs & 82.1 & 82.9 \\ 
    + Depth (24 Layer) & 82.1 & 83.3 \\
    \hline
    \end{tabular}
    \caption{
    More stabilized training of DeiT models with our patch diversification promoted losses.
    }
    \label{tab:ablate_deit_training}
\end{table}

\section{Related Works and Discussions}

\paragraph{Transformers for image classification}
Transformers~\citep{vaswani2017attention} have achieved great success in natural language understanding~\citep{devlin2018bert, radford2019language}, which motivates recent works in adapting transformers to computer vision.
For example, iGPT~\citep{chen2020generative} views an image as a long sequences of pixels and successfully trains transformers to generate realistic images;
\citet{dosovitskiy2020vit} splits each image into a sequence of patches
and achieves competitive performance on challenging ImageNet when pretraining on a large amount of data; 
\citet{touvron2020training} leverages knowledge distillation to improve the training efficacy of vision transformers and achieves better speed vs. accuracy on ImageNet compared to EfficientNets. 

Recently, a variety of works focus on improving vision transformers by 
introducing additional convolutional layers to take advantage of the benefits of inductive bias of CNNs, e.g., 
\citep{han2021transformer, liu2021swin, wu2021cvt, zhou2021deepvit, jiang2021token, touvron2021going, valanarasu2021medical, arnab2021vivit, xie2021sovit, yuan2021tokens}.
Among them, 
LV-ViT~\citep{jiang2021token} is most related to our work, 
LV-ViT~\citet{jiang2021token} also introduces patch-wise auxiliary loss for training, which is equivalent to our \emph{patch-wise mixing loss}; 
LV-ViT is a concurrent work of our submission.
Compared to LV-ViT, our method focuses on improving the patch diversity of vision transformers, which is motivated from a very different perspective. 

\paragraph{Diverse representations in CNNs}
In CNNs, the diverse representation often refers to feature diversity across different channels \citep[e.g.][]{lee2018diverse, liu2018learning}.
In vision transformers, patches could be viewed as feature vectors from different spatial locations in CNNs. 
CNNs' feature maps at different spatial locations are naturally diversified as only local operators are used. 
Transformers, on the other hand, use global attention to fuse features across the different locations, and tends to learn similar presentations without regularization.


\paragraph{Limitations and negative societal impacts}
For the limitation part, 
in this work, we mainly focus on vision transformer. Therefore, our method can only be applied to transformer models with images as input. 
Currently, we only focus on supervised learning problems, and do not study unsupervised or semi-supervised learning. 
Although our work has positive impact for the research community, we also have potential negative social impacts.
Once we open-source our model checkpoints and code, we will have no control of anyone who can get access to the code and model checkpoints.

\section{Conclusion}
In this paper, we 
encourage diversified patch representations
 when training image transformers. 
We address the problem by 
proposing three losses. 
Empirically, without changing the transformer model structure, 
by making patch representations diverse, 
we are able to train larger, deeper models and obtain better performance on image classification tasks.
Applying our pretrained model to semantic segmentation tasks, we obtain SOTA results on two popular datasets, ADE20K and Cityscapes.

For future works,
we plan to study how to encourage diversified patch representations for different tasks. 
We will also incorporate the proposed loss into self-supervised learning settings, and study if the transformer model can serve as a better self-supervised learner for computer vision tasks when the patch representations are more diverse.

\clearpage
\bibliography{transformers}





\end{document}